# OntologyAligner: Ontology-Aligned Retrieval and Hierarchy-Guided Large Language Model Reranking for Biomedical Ontology Normalization

**Authors**

Jie Song[a], PhD; Zhichuan Xu[a], BS; Ziyu Lu[a], BS; Meng Xiao[a], BS; Cheng Bi[b], PhD; Yuxin Zhang[a], PhD; Xin Zheng[a], PhD; Xiaoran Li[a], BS; Qiongfang Cao[a], PhD; Hao Yang[a], PhD; Bairong Shen[a*], PhD;

**Affiliations:**

[a]Department of Dermatovenereology and Institutes for Systems Genetics, Frontiers Science Center for Disease-related Molecular Network, West China Hospital, Sichuan University, Chengdu 610212, China

[b]College of Health and Intelligent Engineering, Chengdu Medical College, Chengdu, Sichuan 610500, China

[*] Corresponding author: Bairong Shen

Corresponding author E-mail: bairong.shen@scu.edu.cn

Corresponding author Phone: +86 15995854635

## The bigger picture

Biomedical research increasingly depends on connecting information expressed in different forms. Biomedical ontologies provide these shared concept systems: they organize entities such as phenotypes, diseases, and species, together with their synonyms, definitions, and relationships. Ontology normalization maps the many expressions found in biomedical text to these stable concepts, enabling data from clinical records, literature, and research databases to be integrated and analyzed consistently. This mapping is especially challenging when two concepts have similar meanings but differ in specificity, anatomical scope, or position within an ontology hierarchy. OntologyAligner addresses this challenge by combining ontology-aware information retrieval, language-model-based semantic comparison, and explicit use of local concept relationships. The resulting framework illustrates how structured biomedical knowledge can guide language models toward more reliable and traceable text-to-concept mapping, supporting downstream phenotyping, knowledge integration, and clinical decision support.

## Highlights

- PhenoNormBench unifies seven HPO datasets containing 13,390 samples
- A three-stage algorithm combines ontology-aligned retrieval, LLM reranking, and hierarchy-guided refinement.
- OntologyAligner achieves state-of-the-art performance across four biomedical ontologies

## Summary

Biomedical ontology normalization maps free-text expressions to standardized concepts, enabling consistent integration and analysis of biomedical data. This task remains challenging because lexical variation and subtle distinctions among hierarchically related concepts can obscure concept boundaries. We present OntologyAligner, a three-stage framework that combines ontology-aligned retrieval, large language model candidate reranking, and selective hierarchy-guided refinement. We also construct PhenoNormBench, a unified benchmark comprising 13,390 samples from seven Human Phenotype Ontology datasets. OntologyAligner achieved state-of-the-art performance on HPO normalization, with 88.78% Macro Top-1 Accuracy and 86.75% Micro Top-1 Accuracy, exceeding the strongest baseline by 4.85 and 5.07 percentage points, respectively. Ablation analyses showed complementary contributions from all three stages, and sensitivity analyses demonstrated stability across candidate-set sizes and model backbones. Applications to MONDO, MEDIC, and NCBITaxon further established portability to other ontologies. OntologyAligner offers a generalizable framework for accurate mapping of biomedical text to structured ontology concepts. PhenoNormBench and the code are publicly available at https://github.com/zhelishisongjie/OntologyAligner.

## INTRODUCTION

Biomedical literature and clinical records contain extensive knowledge about entities such as diseases, drugs, genes, and phenotypes, yet these entities often appear in diverse and nonstandard natural language forms. Enabling data integration [1] and interoperability, clinical decision support [2-6], and large-scale computational analysis requires the reliable mapping of free text to stable ontology concepts, a process known as ontology normalization. Among biomedical ontologies, the Human Phenotype Ontology (HPO) [7] is a widely used knowledge resource for clinical phenotype description and computational analysis. HPO normalization presents several challenges. A single HPO concept often corresponds to multiple lexical variants, and some phenotype mentions have minimal lexical overlap with their synonyms, leading lexical matching methods to miss many valid mappings. In addition, parent, child, and sibling concepts often differ only subtly in anatomical site, clinical characteristics, or semantic granularity, requiring models to perform fine-grained semantic discrimination.

Early dictionary- and rule-based methods are simple and transparent, with traceable decision processes [8, 9]. Their coverage is restricted to labels and synonyms included in the ontology, limiting generalization to unrecorded lexical variants. Neural networks and dense retrieval can accommodate a broader range of semantic variants [10, 11]. However, general-purpose embedding spaces remain misaligned with the concept boundaries of the target ontology, and a single similarity score provides limited discrimination among highly similar candidate concepts. In recent years, large language models (LLMs) have been applied to this task [12]. Retrieval-augmented generation incorporates external knowledge, such as term definitions, enabling models to combine textual context with reasoning for concept disambiguation [13]. Existing methods primarily reason from the textual semantics of candidate concepts, with limited explicit use of ontology hierarchies, and their candidate retrieval spaces remain misaligned with the target ontology.

To address these challenges, we propose OntologyAligner, a three-stage framework integrating ontology-aligned retrieval, semantic candidate reranking, and selective hierarchy-guided refinement. First, Ontology-Aligned Retrieval (OAR) learns a target-ontology-specific projection for dense retrieval, mapping input phrases and ontology term variants into an aligned representation space to retrieve candidate concepts. Next, LLM Candidate Reranking (LCR) performs fine-grained semantic ranking over the complete candidate set by jointly considering each candidate's preferred label, synonyms, and definition. Finally, when the Top-1 results from OAR and LCR differ and the relevant high-ranking candidates exhibit local parent-child or sibling relationships, Hierarchy-Guided Refinement (HGR) incorporates explicit hierarchical evidence and reranks the candidates. The three modules work synergistically to support candidate retrieval, semantic discrimination, and the resolution of local hierarchical conflicts. We also integrated seven HPO datasets from diverse sources to construct a unified evaluation benchmark, PhenoNormBench, and evaluated the cross-ontology portability of OntologyAligner on MONDO, MEDIC, and NCBITaxon. In summary, our contributions are as follows:

- We propose OAR, which learns a target-ontology-specific projection through concept-level multi-positive InfoNCE loss, using same-concept term variants as positives and neighboring variants from other concepts as hard negatives to improve candidate retrieval.
- We propose HGR, a selective refinement mechanism that uses explicit local hierarchical evidence to distinguish semantically similar candidates when OAR and LCR produce different Top-1 predictions involving parent-child or sibling relationships.
- We construct PhenoNormBench by integrating seven HPO datasets and establish a unified evaluation protocol for ontology versions and acceptable ID normalization. Comprehensive baseline comparisons, ablation studies, and sensitivity analyses demonstrate OntologyAligner's robustness and state-of-the-art performance, while experiments on MONDO, MEDIC, and NCBITaxon establish its cross-ontology portability.

## RESULTS

### OntologyAligner improves HPO normalization in PhenoNormBench

OntologyAligner achieved a Micro Top-1 Accuracy of 86.75% (95% CI: 86.18 to 87.32) and an equal-weighted Macro Top-1 Accuracy of 88.78% across the seven datasets (95% CI: 88.23 to 89.33) (**Figure 1**). The strongest baseline, RAG-HPO-TE3L, achieved Macro and Micro Accuracy values of 83.93% and 81.68%, respectively; OntologyAligner improved these results by absolute differences of 4.85 and 5.07 in Macro and Micro Accuracy, respectively. Relative to RAG-HPO-original, OntologyAligner achieved absolute differences of 6.88 and 7.09 in Macro and Micro Accuracy, respectively, with larger gains over the three conventional LLM-free baselines. All five overall baseline comparisons reached statistical significance in both the Macro and Micro panels (P_adj < 0.001).

OntologyAligner achieved the highest Top-1 Accuracy on all seven datasets: 93.30% on FGDD, 81.66% on CSC, 97.16% on GeneReviews-10, 88.94% on BC8-T3, 80.92% on GSC2017, 85.74% on GSC2024, and 93.76% on ID-68 (**Supplementary Figure S1**). Relative to RAG-HPO-TE3L, the absolute improvements across datasets ranged from 4.15% to 7.04%, with the smallest improvement on ID-68 and the largest on BC8-T3. All dataset-level comparisons with the other four baselines were also statistically significant (P_adj < 0.001).

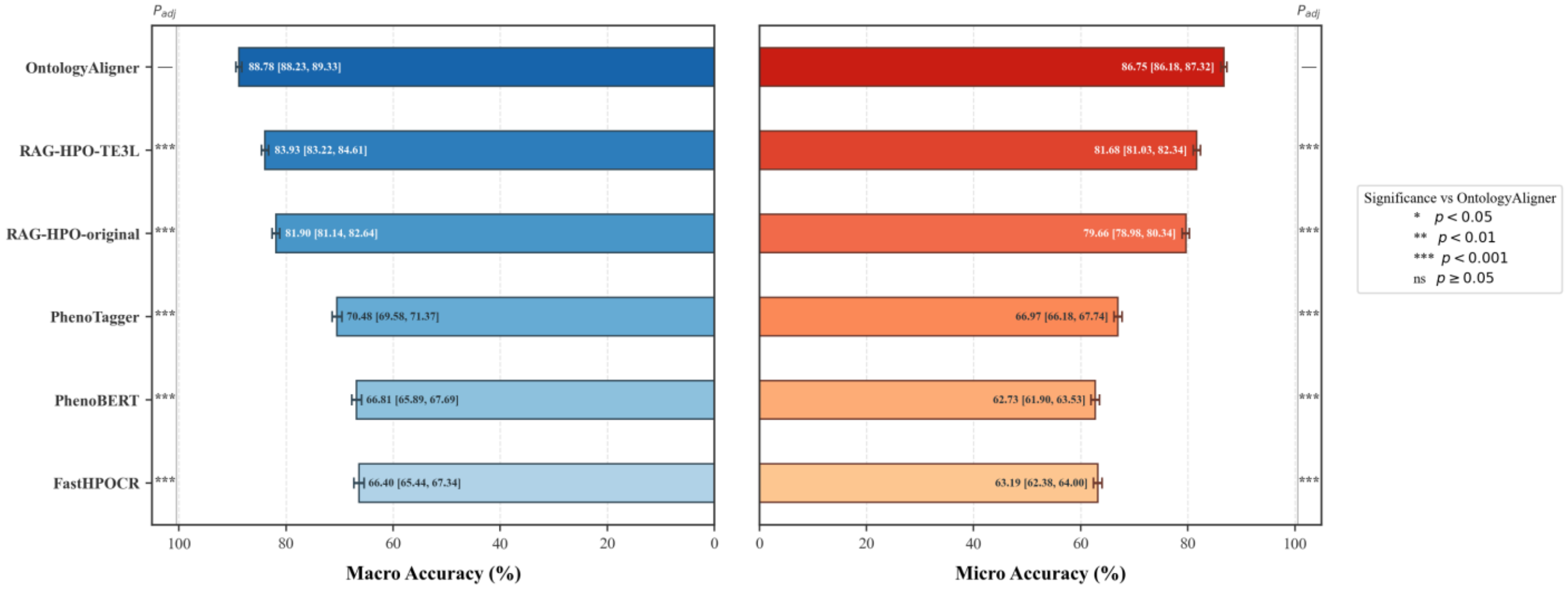

**Figure 1. Overall Top-1 accuracy across seven benchmark datasets.** Macro Accuracy is the unweighted mean of seven dataset-level accuracies; Micro Accuracy is pooled over 13,390 samples. Macro uses paired dataset-stratified bootstrap and Micro uses pooled exact McNemar tests; five comparisons per panel use Holm correction. *** P_adj<0.001.

Difficulty-stratified analysis is presented in **Figure 2**. Stratification by phrase length (**Figure 2(a)**) showed Top-1 Accuracy values of 91.94% for one-token phrases, 88.72% for two-token phrases, 88.39% for three-token phrases, and 73.16% for phrases containing at least four tokens. The latter three groups had significantly lower accuracy than the one-token group (all P_adj < 0.001).

Stratification by lexical overlap (**Figure 2(b)**) showed a Top-1 Accuracy of 58.13% for phrases with zero lexical overlap with the target HPO term variants. Accuracy increased with lexical overlap, reaching 74.58%, 84.06%, and 99.92% for the low-, high-, and exact-overlap groups, respectively. Each of the zero-, low-, and high-overlap groups differed significantly from the exact-overlap reference group (all P_adj < 0.001).

Ontology depth showed a nonmonotonic association with accuracy (**Figure 2(c)**). Compared with the reference group of depth 1 to 4, the differences for depths 5 to 6 and 9 to 12 were not statistically significant, whereas the depth 7 to 8 group had significantly lower accuracy (84.59%, P_adj < 0.001).

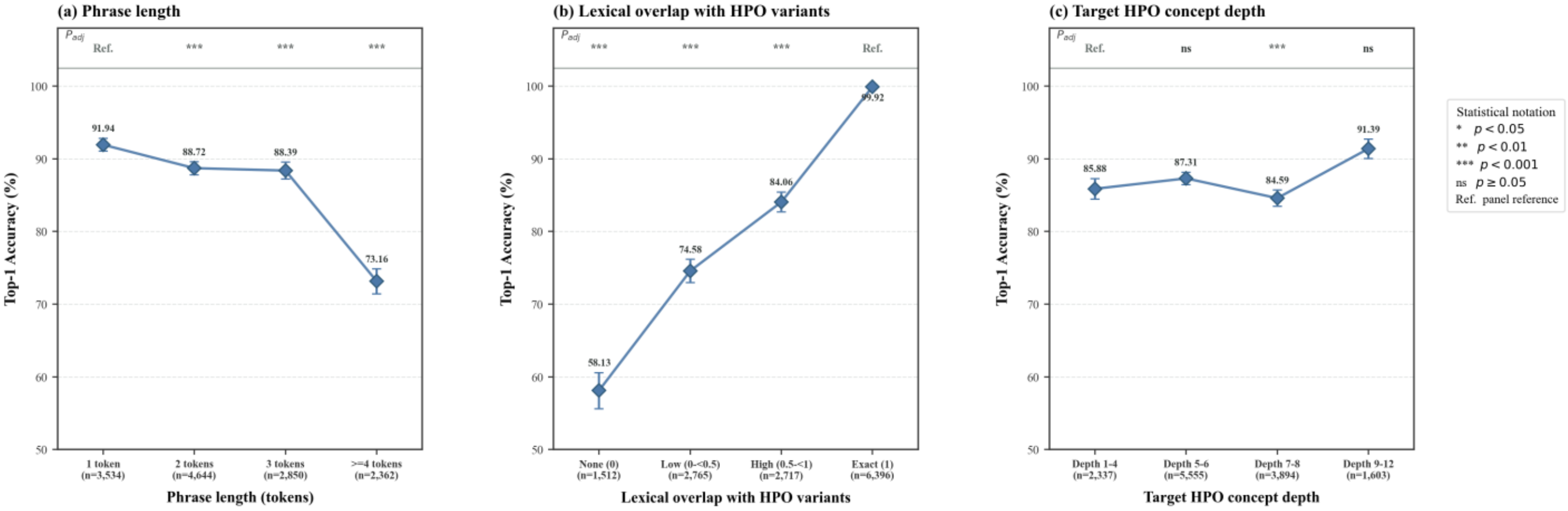

**Figure 2. Performance of OntologyAligner stratified by phenotype-phrase difficulty.** (a) Top-1 Accuracy according to phrase length. (b) Top-1 Accuracy according to the maximum token-set Jaccard overlap between each phrase and its target HPO term variants. (c) Top-1 Accuracy according to target HPO concept depth. Statistical comparisons use two-sided Cochran Mantel Haenszel tests stratified by benchmark, with 1 token, exact lexical overlap, and depth 4 as the respective reference strata. Three comparisons within each panel are adjusted using the Holm method. *** P_adj<0.001; ns, P_adj≥0.05.

### Component contributions and robustness analyses

Component ablation results for all 13,390 samples are shown in **Figure 3(a)**. Across all samples, the Macro/Micro Accuracy values for Raw Retrieval Top-1, OAR, OAR + LCR, and the full OAR + LCR + HGR model were 83.07%/80.82%, 84.71%/82.58%, 88.56%/86.65%, and 88.78%/86.75%, respectively. The improvements from OAR and LCR relative to the preceding conditions were statistically significant (P_adj < 0.001). Adding HGR increased Macro and Micro Accuracy by absolute differences of 0.22 and 0.10, respectively (P_adj < 0.05). Raw Retrieval + LCR + HGR (w/o OAR) achieved Macro/Micro Accuracy values of 87.90%/85.97%, representing absolute differences of 0.88 and 0.78 compared with the full model (P_adj < 0.001).

**Supplementary Figure S2** presents two cases in which incorrect predictions were converted into correct predictions. In the case of “homocysteine that were persistently elevated,” LCR ranked HP:0010919 (Abnormal circulating homocysteine concentration) first. After HGR adjustment, its descendant concept and the reference standard concept, HP:0002160 (Hyperhomocystinemia), were promoted to the first position. In the case of “lack of sleep features,” LCR ranked HP:0002360 (Sleep disturbance) first. The local ontology relationships showed that the reference standard concept HP:0100785 (Insomnia) and the candidate concept HP:4000064 (Poor sleep) both had HP:0002360 as their direct parent. After HGR adjustment, HP:0100785 (Insomnia) was promoted to the first position.

To ensure equal representation of the seven datasets while maintaining computational feasibility, we constructed a balanced subset comprising 300 samples from each dataset. Macro Accuracy and Micro Accuracy were identical on this balanced subset; therefore, Top-1 Accuracy is reported throughout the following analysis. The sensitivity analysis of candidate-set size is shown in **Figure 3(b)**. At $k = 1$, the accuracy was 83.48%; for $k \geq 3$, accuracy ranged from 87.48% to 87.95%. Using $k = 20$ as the reference, the accuracy at $k = 1$ was significantly lower ($P_{adj} < 0.001$), while the differences for $k = 3$, 5, and 10 were not statistically significant.

The comparison of different embedding backbones is shown in **Figure 3(c)**. TE3L achieved an accuracy of 87.67%. TE3-small and BioLORD each achieved an accuracy of 87.10%, with no statistically significant difference from TE3L. BioBERT, PubMedBERT, and ClinicalBERT achieved accuracies of 82.43%, 82.05%, and 81.57%, respectively, all significantly lower than TE3L ($P_{adj} < 0.001$).

The comparison of different LLM backbones is shown in **Figure 3(d)**. Claude Opus 5 achieved an accuracy of 88.62%, representing an absolute increase of 0.95 over GPT-5.6-sol, which achieved 87.67% ($P_{adj} < 0.01$). DeepSeek V4 Flash and DeepSeek V4 Pro achieved accuracies of 87.52% and 87.24%, respectively; both differences from GPT-5.6-sol were statistically nonsignificant.

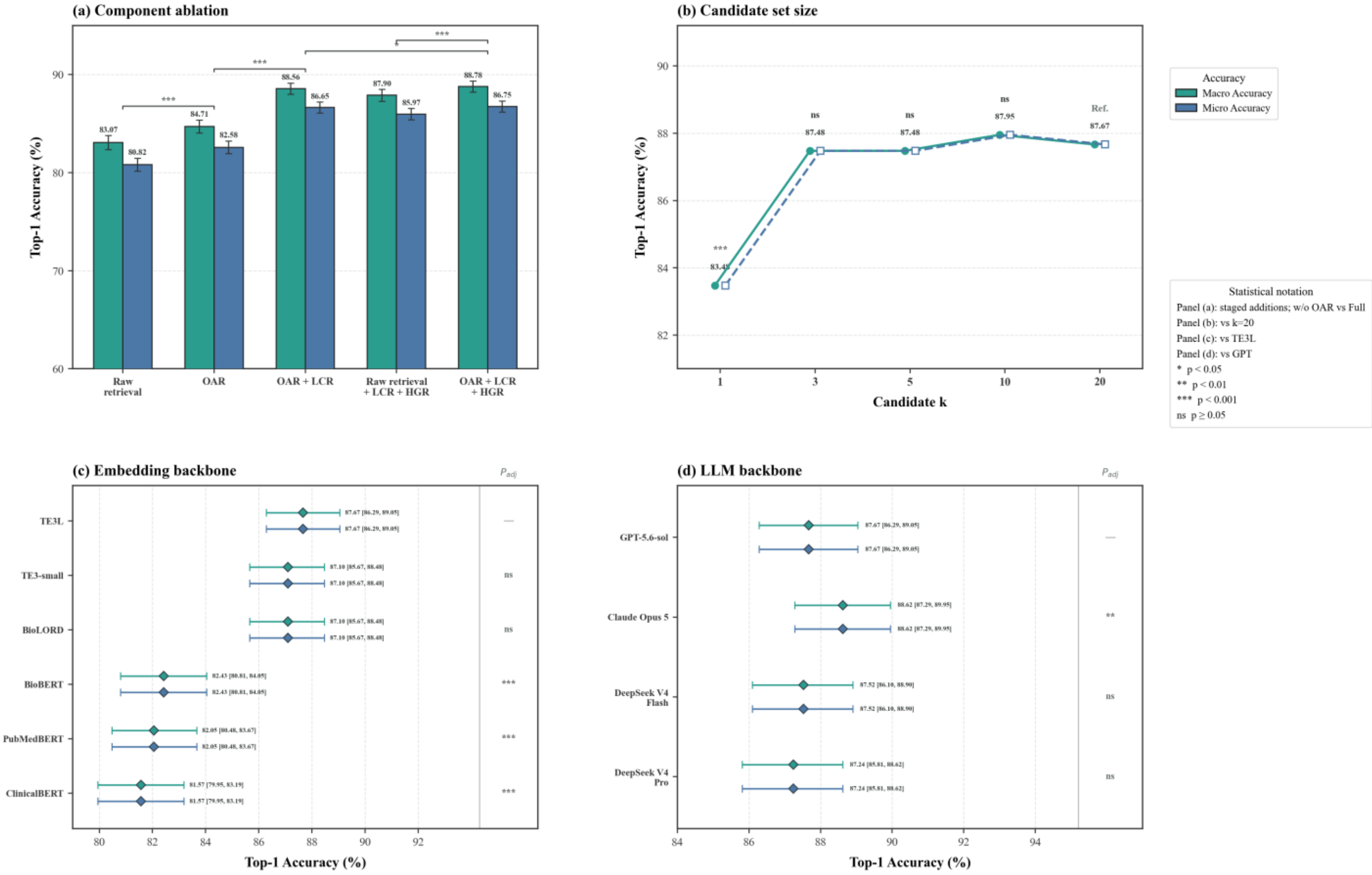

**Figure 3. Ablation and sensitivity analyses of OntologyAligner.** (a) Component ablation on all 13,390 samples, comparing Raw retrieval, OAR, OAR + LCR, Raw retrieval + LCR + HGR (w/o OAR), and the full OAR + LCR + HGR. (b) Candidate-set-size analysis on the balanced 2,100-sample subset, with k=20 as the reference condition. (c) Embedding-backbone analysis on the balanced 2,100-sample subset, with TE3L as the reference condition. (d) LLM-backbone analysis on the balanced 2,100-sample subset, with GPT-5.6-sol as the reference condition. Significance symbols indicate Holm-adjusted paired exact McNemar tests: * P_adj<0.05, ** P_adj<0.01, *** P_adj<0.001, and ns P_adj>=0.05.

### Generalization to other biomedical ontologies

We further evaluated OntologyAligner on three biomedical ontology normalization tasks, training a separate projection and constructing a dedicated vector index for each target ontology. The three experiments covered two disease ontologies and one biological taxonomy: FGDD Disease mapped to MONDO, the NCBI Disease Corpus mapped to MEDIC, and species names from S800 mapped to NCBITaxon. OntologyAligner achieved Top-1 Accuracy values of 97.66%, 92.19%, and 94.13% on the three tasks, respectively (**Table 1**).

The performance of the comparison methods was taken directly from the results reported by BELB and BELHD. For mapping the NCBI Disease Corpus to MEDIC, the reported comparison methods achieved Top-1 Accuracy values ranging from 79.90% to 87.60%, with BELHD achieving the highest accuracy at 87.60%. OntologyAligner achieved an accuracy of 92.19%, representing an absolute increase of 4.59 over BELHD. For mapping S800 to NCBITaxon, the reported comparison methods achieved Top-1 Accuracy values ranging from 78.62% to 89.96%, with GenBioEL+HD achieving the highest accuracy at 89.96%. OntologyAligner achieved an accuracy of 94.13%, representing an absolute increase of 4.17 over GenBioEL+HD.

**Table 1. Results on other biomedical ontologies.**

| **Dataset** | **n** | **Ontology** | **OntologyAligner** | **arboEL** [14] | **BioSyn** [15] | **GenBioEL** [16] | **GenBioEL+HD** [17] | **BELHD** [17] |
|---|---|---|---|---|---|---|---|---|
| FGDD Disease | 256 | MONDO | 97.66 | - | - | - | - | - |
| NCBI Disease Corpus | 960 | MEDIC | 92.19 | 80.00 | 79.90 | 82.71 | 83.02 | 87.60 |
| S800 | 767 | NCBITaxon | 94.13 | 78.62 | 82.79 | 88.27 | 89.96 | 84.35 |

## DISCUSSION

Ontology normalization provides the foundation for connecting heterogeneous biomedical text with structured knowledge resources. In this study, OntologyAligner achieved a Macro Top-1 Accuracy of 88.78% and a Micro Top-1 Accuracy of 86.75% across 13,390 samples from seven HPO datasets, representing absolute improvements of 4.85 and 5.07 percentage points, respectively, over the strongest baseline.

The ablation experiments revealed complementary contributions from the three stages. By learning a target-ontology-specific representation space, OAR increased Micro Accuracy from 80.82% to 82.58%, indicating that ontology-space alignment can improve candidate retrieval quality. LCR further increased Micro Accuracy to 86.65% and contributed the largest improvement to overall performance. The full model also outperformed the configuration without OAR, demonstrating that high-quality candidate retrieval provides a more reliable foundation for downstream semantic reranking.

HGR provided a smaller yet statistically supported improvement, consistent with its selective triggering mechanism. Specifically, HGR focuses on local hierarchical conflicts that arise when the OAR and LCR predictions differ and the competing candidates exhibit parent-child or sibling relationships. Explicit hierarchical evidence helps the model assess concept scope and specificity, while the homocysteine- and sleep-related cases demonstrate its ability to resolve local structural ambiguities. The primary value of HGR therefore lies in its targeted correction of a limited number of difficult samples and its provision of auditable ontology-relation evidence for the final prediction.

The difficulty-stratified analysis showed that lexical divergence and phrase complexity were major sources. Accuracy was 58.13% for phrases with zero lexical overlap and 73.16% for phrases containing four or more tokens, whereas phrases with exact lexical matches achieved near-ceiling performance. These findings indicate that paraphrastic expressions, compositional phenotype descriptions, and phrases containing multiple clinical modifiers remain challenging for ontology normalization.

The sensitivity analyses further demonstrated the framework's stability across multiple implementation configurations. Candidate sets containing three or more concepts achieved accuracy comparable to that obtained with 20 candidates, suggesting that smaller candidate sets could reduce LLM input length, inference latency, and computational cost. After ontology alignment, TE3-small and BioLORD performed comparably to TE3L, while multiple LLM backbones produced broadly consistent results, indicating that the three-stage pipeline can accommodate different embedding and LLM backbones. In practical deployment, the model and candidate-set size can be selected according to requirements for accuracy, cost, latency, and data governance.

The experiments on MONDO, MEDIC, and NCBITaxon extended the evaluation to disease and taxonomic entity normalization. After training an ontology-specific projection and constructing a dedicated index for each target ontology, OntologyAligner achieved high Top-1 Accuracy across all three tasks, supporting the framework's portability across different biomedical entity types.

From an application perspective, OntologyAligner organizes candidate retrieval, semantic ranking, and hierarchical refinement into independent and traceable decision stages. Researchers can examine candidate concepts, terminological evidence, and local ontology relationships separately, enabling concept-level error analysis and quality control. This modular and auditable workflow can support phenotype annotation, cohort construction, literature curation, and biomedical knowledge integration, while providing a structured basis for error analysis and human review in future normalization systems.

**Limitations**

This study has several limitations. First, the evaluation focused on pre-extracted, isolated biomedical phrases and therefore primarily measured the performance of the ontology normalization module at the phrase level. Accordingly, the current findings apply mainly to standard phrase-level evaluation settings, while performance across the complete clinical text-processing pipeline requires end-to-end validation. Second, HGR currently uses parent-child and sibling relationships to resolve local concept conflicts, primarily distinguishing semantically similar concepts according to their specificity, semantic scope, and hierarchical position. Biomedical ontologies also contain other relation types, such as part_of and has_modifier, and structural information from longer hierarchical paths may provide additional evidence for concept discrimination. Third, the current experiments primarily covered English biomedical expressions, so the applicability of the framework to other languages requires further evaluation. Different languages exhibit systematic differences in medical terminology, word order, abbreviation patterns, and modifier structures; cross-lingual normalization is also influenced by the quality of terminology translation and ontology localization resources.

**Future work**

Future work should extend OntologyAligner to a complete clinical text-processing pipeline that jointly performs entity recognition, contextual understanding, and ontology normalization. The ontology-structure information used by HGR could also be expanded by investigating relations such as part_of and has_modifier, together with longer hierarchical paths, beyond the current use of parent–child and sibling relationships. Ontology-specific evidence organization and reasoning rules for different relation types may improve the discrimination of complex concept boundaries. Future studies could further investigate OntologyAligner in multilingual biomedical text by integrating multilingual terminology resources, ontology localization mappings, and manually annotated corpora to train corresponding retrieval projections and reranking modules. A unified benchmark covering different languages, region-specific terminology, and cross-lingual expressions would support systematic evaluation of multilingual transfer performance.

**Conclusions**

This study presents OntologyAligner, a three-stage biomedical ontology normalization framework that integrates ontology-aligned retrieval, LLM-based candidate reranking, and selective hierarchy-guided refinement. OAR aligns queries with the terminology space of the target ontology, LCR performs fine-grained semantic discrimination by jointly considering candidate labels, synonyms, and definitions, and HGR resolves hierarchical conflicts using local parent–child and sibling relationships. On PhenoNormBench, which contains 13,390 samples, OntologyAligner achieved a Macro Top-1 Accuracy of 88.78% and a Micro Top-1 Accuracy of 86.75%, representing absolute improvements of 4.85 and 5.07 percentage points, respectively, over the strongest baseline. Ablation experiments showed that OAR, LCR, and HGR each made independent contributions to overall performance, with LCR providing the largest improvement and HGR offering a statistically supported additional correction for a small number of local hierarchical ambiguities. Experiments on MONDO, MEDIC, and NCBITaxon further showed that, after training an ontology-specific retrieval projection and constructing a dedicated index for each target ontology, the workflow could be applied to biomedical entity normalization tasks involving phenotypes, diseases, and biological taxa. Overall, OntologyAligner provides a modular, extensible, and locally traceable solution for mapping biomedical text to ontology concepts, supporting phenotype integration, literature curation, cohort construction, and downstream biomedical computational analyses.

## METHODS

### Unified HPO normalization benchmark

Existing HPO normalization datasets vary in ontology versions and annotation standards and have yet to be systematically integrated, complicating consistent evaluation across methods. To address this, we standardized and consolidated seven phenotype normalization datasets to construct PhenoNormBench. We first mapped all concept identifiers to the latest HPO release (2026-06-23). To reduce evaluation errors arising from ontology version differences, PhenoNormBench defines Accepted HPO IDs for each concept. These include the Standard HPO ID and all officially documented alt_ids, which are historical identifiers retained by HPO for the same concept; a prediction matching any Accepted HPO ID is considered correct. We then used individual phenotype phrases as the basic evaluation units and converted all datasets into a standardized structure containing the phrase, Standard HPO ID, and Accepted HPO IDs. **Table 2** presents the sources and sizes of the datasets. The seven datasets contain 13,390 samples in total, of which 7,355 have multiple Accepted HPO IDs.

**Table 2.** Composition of PhenoNormBench. Sample counts are reported for the seven constituent HPO normalization datasets.

| **Dataset** | **Samples** |
|---|---|
| FGDD [18] | 1,866 |
| CSC [13] | 1,783 |
| GeneReviews-10 [19] | 352 |
| BC8-T3 [20] | 2,840 |
| GSC2017 [8] | 2,773 |
| GSC2024 [9] | 2,910 |
| ID-68 [11] | 866 |

### OntologyAligner

OntologyAligner is a three-stage framework for ontology normalization. It comprises Ontology-Aligned Retrieval (OAR), LLM Candidate Reranking (LCR), and Hierarchy-Guided Refinement (HGR), organizing ontology-aligned retrieval, semantic assessment, and hierarchical disambiguation into three consecutive stages. Given a phenotype phrase, OAR first retrieves 20 candidate concepts from an ontology-aligned representation space. LCR then semantically reranks the candidates based on their terminological information, and HGR selectively refines the ranking by incorporating local ontology hierarchy information when its triggering conditions are met. The framework ultimately outputs the top-ranked HPO ID. An overview of the OntologyAligner framework is shown in **Figure 4.**

***Ontology-Aligned Retrieval***

OAR aims to transform a general-purpose embedding space into a retrieval space that better aligns with the target ontology. We treat the preferred label and synonyms of each HPO concept as concept variants and use the text-embedding-3-large (TE3L) [21] model to obtain 3,072-dimensional initial embeddings. Each initial embedding is passed through a shared bias-free linear projection followed by L2 normalization, yielding an ontology-aligned embedding:

$$\mathbf{z}_x = \frac{\mathbf{W}\mathbf{e}_x}{\|\mathbf{W}\mathbf{e}_x\|_2}, \tag{1}$$

Where $\mathbf{e}_x$ is the initial embedding of term variant $x$, and $\mathbf{W} \in \mathbb{R}^{3072\times3072}$ is a projection matrix initialized as the identity matrix.

During training, each term variant serves as an anchor, with the remaining variants of the same concept used as positive samples. To mine hard negative samples, we retrieve the anchor's 50 nearest variants in the initial embedding space, group them by concept, and score each concept by its maximum variant-level cosine similarity. The 10 highest-scoring neighboring concepts are selected as hard negatives. The matching score between anchor $x_i$ and concept $c$ is defined as:

$$S(x_i, c) = \max_{v\in V_c} \mathbf{z}_{x_i}^{\top} \mathbf{z}_v, \tag{2}$$

where $V_c$ is the set of term variants for concept $c$, with the anchor excluded when scoring its own concept.

The projection matrix is trained using the Information Noise-Contrastive Estimation (InfoNCE) loss [22]:

$$\mathcal{L} = -\frac{1}{|\mathcal{A}|}\sum_{i\in\mathcal{A}} \log \frac{\exp\left(\frac{S(x_i, c_i)}{\tau}\right)}{\exp\left(\frac{S(x_i, c_i)}{\tau}\right) + \sum_{c\in N_i} \exp\left(\frac{S(x_i, c)}{\tau}\right)} + \lambda \text{ mean } [(\mathbf{W}-\mathbf{I})^2], \tag{3}$$

where $\mathcal{A}$ is the anchor set, $c_i$ is the positive concept associated with anchor $x_i$, and $N_i$ is its set of hard negative concepts. The contrastive term increases $S(x_i, c_i)$ relative to $S(x_i, c)$ for $c \in N_i$, thereby pulling variants of the same concept closer to the anchor and pushing confusable hard-negative concepts farther away in the projected embedding space. The identity-regularization term constrains $W$ to remain close to the identity matrix, limiting excessive distortion of the original embedding space. The temperature is set to $\tau = 0.05$, $I \in \mathbb{R}^{3072\times3072}$ is the identity matrix, and $\lambda$ is the regularization coefficient.

After training, we use text-embedding-3-large to encode 44,814 concept variants (preferred labels and synonyms) from 19,836 HPO concepts. The resulting embeddings are mapped through the trained projection matrix $W^*$, L2-normalized, and stored in a Chroma vector database. During inference, an input phenotype phrase undergoes the same encoding, projection, and normalization procedure. Its embedding is then compared with the embeddings of all HPO concept variants in the vector database using cosine similarity. The 20 HPO concepts with the highest cosine-similarity scores are returned as the candidate set for LCR.

***LLM Candidate Reranking***

OAR retrieves semantically relevant candidates, while fine-grained distinctions among HPO concepts with similar meanings but different scopes require additional semantic assessment. LCR uses candidate terminology to perform fine-grained semantic comparison and reranking. It employs gpt-5.6-sol [23] as the LLM backbone with the temperature set to 0. Given the original phenotype phrase and each candidate's preferred label, synonyms, and definition, LCR compares and fully ranks the 20 candidates returned by OAR, selecting the most specific concept that is semantically equivalent to and fully captures the input phrase. The resulting ranking is then passed to HGR. The complete prompt used in LCR is provided in **Supplementary Table S2**.

***Hierarchy-Guided Refinement***

HPO concepts connected by parent-child or sibling relationships are often semantically similar, making their hierarchical boundaries difficult to distinguish using preferred labels, synonyms, and definitions alone.

When the Top-1 predictions from OAR and LCR differ, HGR compares the OAR Top-1 concept with each of the other candidates in the LCR Top-3. HGR is triggered when any pair has a parent-child or sibling relationship. The relevant concepts, explicit is_a relations, and definitions are then organized into a local graph context and provided to the LLM to rerank candidates. Cases without such relationships retain the LCR ranking. The complete prompt used in HGR is provided in **Supplementary Table S2**

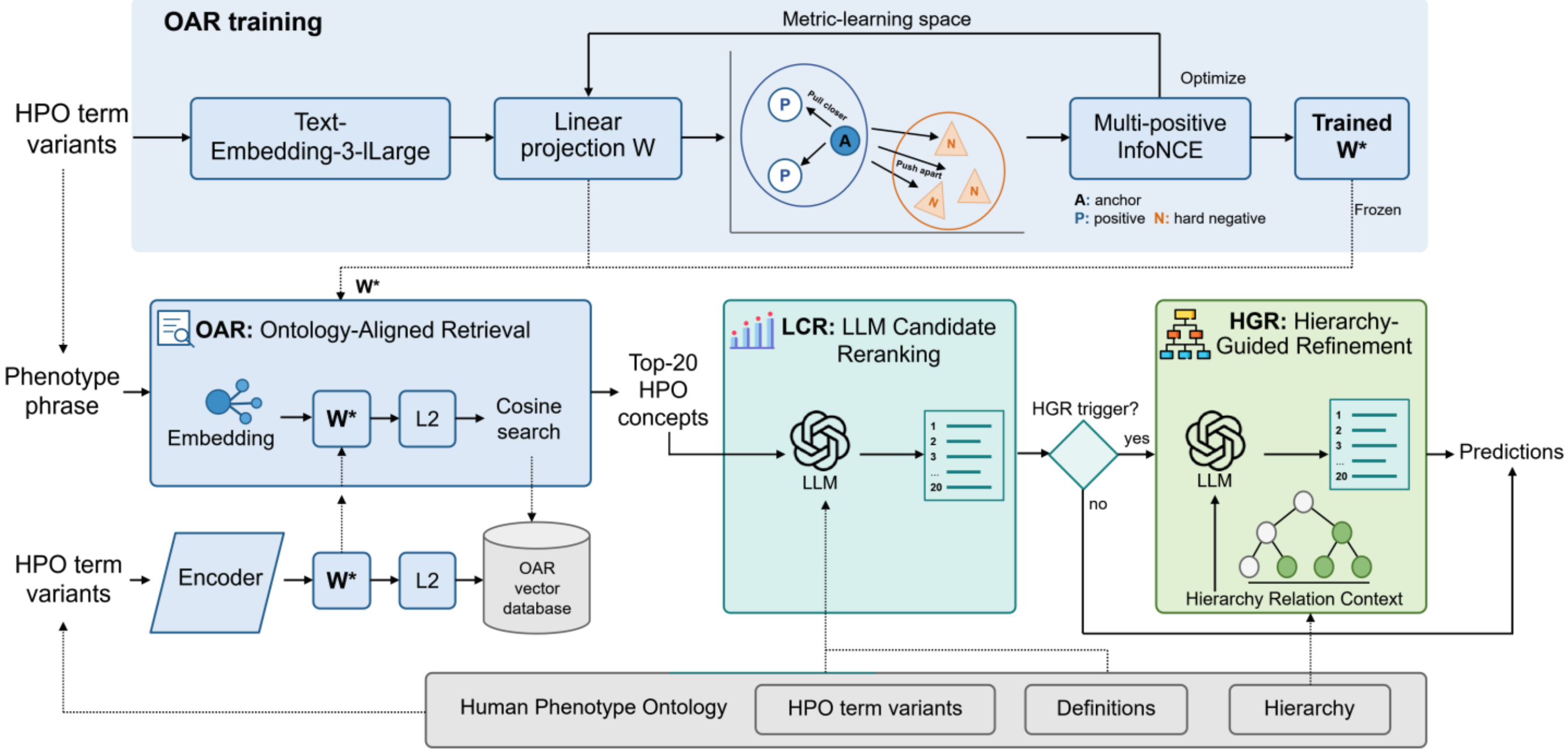

**Figure 4. Overview of the OntologyAligner framework.** OAR training uses concept-level multi-positive InfoNCE to learn a bias-free linear projection. At inference, OAR applies the frozen projection to retrieve the Top-20 HPO concepts from an ontology-aligned vector database, and LCR reranks the candidates using term variants and definitions. When the OAR and LCR Top-1 predictions differ and the relevant candidates exhibit local parent-child or sibling relations, HGR uses this hierarchy context to refine the LCR ranking and generate the final prediction.

## Experimental design and evaluation

### *HPO normalization*

We compare OntologyAligner with FastHPOCR [9], PhenoBERT [11], PhenoTagger [10], RAG-HPO-original [13], and RAG-HPO-TE3L [13] on PhenoNormBench. This study evaluates only the phenotype normalization; phenotype extraction in the original pipeline is outside the scope of evaluation.

FastHPOCR rebuilds its rule-based index using the current HPO release. PhenoBERT and PhenoTagger are evaluated using their original publicly released models and decision thresholds. RAG-HPO-original constructs its vector index with the bge-small-en-v1.5 model used in the original study. RAG-HPO-TE3L constructs the index with the general-purpose text-embedding-3-large model. For both RAG-HPO variants, the LLM backbone in the HPO assignment stage is upgraded from LLaMA-3.1 70B used in the original study to gpt-5.6-sol. Both variants use the same Top-20 candidate set and LLM assignment prompt.

As supplementary analyses, all methods are additionally evaluated on each of the seven constituent datasets to assess performance consistency across data sources. OntologyAligner is further evaluated across difficulty strata defined by phrase length, maximum token-set Jaccard overlap with the target HPO term variants, and target HPO concept depth.

***Ablation and sensitivity analyses***

To quantify the contribution of each stage, we conducted stage-wise ablation on the full set of 13,390 samples, comparing five configurations: raw retrieval, OAR, OAR+LCR, raw retrieval+LCR+HGR, and the full OAR+LCR+HGR framework.

Sensitivity analyses were conducted on a balanced subset of 2,100 samples, created by randomly sampling 300 samples from each dataset with a sampling seed of 42, thereby assigning equal weight to all seven datasets. The candidate-size analysis varied the number of candidates passed to subsequent stages (k) among 1, 3, 5, 10, and 20. The embedding analysis compared text-embedding-3-large, text-embedding-3-small [21], BioLORD [24], BioBERT [25], PubMedBERT [26], and ClinicalBERT [27]. For each embedding backbone, a separate OAR projection was trained using its native embedding dimensionality, while the downstream LCR and HGR settings were held constant. The LLM analysis used candidate sets generated with TE3L and replaced the LLM backbone in LCR and HGR with Claude Opus 5 [28], DeepSeek V4 Flash [29], or DeepSeek V4 Pro [29]. All LLMs are accessed through API requests. The specific versions of LLMs and embedding models are provided in **Supplementary Table S1**.

***Generalization to other ontologies***

To evaluate the transferability of OntologyAligner beyond HPO, we extended it to MONDO (version 2026-07-06) [30], MEDIC (version 2026-07-30) [31], and NCBITaxon (version 2026-08-06) [32]. For each ontology, we trained an ontology-specific OAR projection, constructed a dedicated vector index. All remaining settings were held constant, including Top-20 candidate retrieval, gpt-5.6-sol, LCR semantic reranking, and the HGR triggering and reranking procedures. The MONDO experiment used the FGDD Disease dataset [18], comprising 256 rare disease phrases from 509 documents. The MEDIC experiment used the NCBI Disease Corpus [33], comprising 960 disease phrases from 100 documents. The NCBITaxon experiment used the S800 dataset [34], comprising 767 organism phrases from 125 documents.

***Metrics and statistical analysis***

The primary evaluation metric was Top-1 Accuracy. Dataset-level Accuracy was calculated separately for each dataset. Macro Accuracy was defined as the unweighted mean accuracy across the seven datasets, and Micro Accuracy as the overall accuracy across all pooled samples.

The 95% CIs for accuracy were estimated using 20,000 percentile bootstrap replicates. Two-sided p values for differences in Micro Accuracy were calculated using the exact McNemar test with Holm correction, whereas those for differences in Macro Accuracy were calculated using a paired, dataset-stratified bootstrap with Holm correction.

Stage-wise ablation was conducted on the full evaluation set. Analyses of candidate-set size, embedding backbone, and LLM backbone were conducted on a fixed balanced subset of 2,100 samples, using k=20, TE3L, and gpt-5.6-sol as the respective reference conditions. All other settings were identical to those of the full OntologyAligner. For these analyses, two-sided p values were calculated using the exact McNemar test with Holm correction. Because the balanced subset contained 300 samples from each dataset, Macro Accuracy and Micro Accuracy were identical.

Difficulty analyses stratified samples by phrase length, maximum token-set Jaccard overlap with the target HPO term variants, and target HPO concept depth. Two-sided p values for group-wise comparisons were calculated using the Cochran Mantel Haenszel test stratified by benchmark, with Holm correction. Cross-ontology experiments reported Top-1 Accuracy only, and comparator performance was taken directly from the results reported in BELB [35] and BELHD [17].

## RESOURCE AVAILABILITY

### Lead contact

Requests for further information and resources should be directed to and will be fulfilled by the lead contact, Bairong Shen (bairong.shen@scu.edu.cn).

### Materials availability

This study did not produce any new biological materials.

### Data and code availability

PhenoNormBench, a benchmark integrating seven standardized phenotypic datasets, and the code used in this study are available at https://github.com/zhelishisongjie/OntologyAligner.

## ACKNOWLEDGMENTS

This work was supported by National Natural Science Foundation of China (Grant No. 32570773 and 32270690), Pioneer and Leading Goose R&D Program of Zhejiang Province (Grant No. 2026C01022). The funder of the study had no role in study design, data collection, data analysis, data interpretation, or writing of the report.

## AUTHOR CONTRIBUTIONS

Jie Song, and Bairong Shen conceptualised the study. Jie Song designed the methodology. Jie Song, Zhichuan Xu, Ziyu Lu, Meng Xiao, Cheng Bi, Yuxin Zhang, Xin Zheng, Xiaoran Li, Qiongfang Cao and Hao Yang curated the datasets. Jie Song implemented the computational analysis workflow, performed the model experiments, analysed the data, and generated the figures and tables. Bairong Shen supervised the study, provided administrative and material support, and acquired funding. Jie Song drafted the initial manuscript; Bairong Shen provided critical revisions. All authors reviewed and approved the final version of the manuscript, had full access to all the data in the study, and had final responsibility for the decision to submit for publication.

## DECLARATION OF INTERESTS

The authors declare no competing interests.

## DECLARATION OF GENERATIVE AI AND AI-ASSISTED TECHNOLOGIES IN THE WRITING PROCESS

During the preparation of this manuscript, the authors used ChatGPT to assist with formatting checks, tense consistency, grammar correction, and language polishing. The authors subsequently reviewed and edited the generated content and assume full responsibility for the final version of the manuscript.

## REFERENCES

1. He M, Song J, Ren S, Zhang Y, Du J, Feng J, et al. FPGDKG 1.0: an integrated facial phenotype-gene-disease knowledge graph for rare disease diagnosis and explanation. IEEE Journal of Biomedical and Health Informatics. 2026.
2. Yang J, Shu L, Duan H, Li H. RDguru: A Conversational Intelligent Agent for Rare Diseases. IEEE Journal of Biomedical and Health Informatics. 2025;29(9):6366-78. doi: 10.1109/JBHI.2024.3464555.
3. Yang J, Shu L, Han M, Pan J, Chen L, Yuan T, et al. RDmaster: A novel phenotype-oriented dialogue system supporting differential diagnosis of rare disease. Computers in Biology and Medicine. 2024 2024/02/01/;169:107924. doi: https://doi.org/10.1016/j.compbiomed.2024.107924.
4. Song J, Xu Z, Xiao M, Bi C, Zhang Y, Zheng X, et al. Initial-Visit Specialty Triage in Rare Diseases Using Large Language Models: Retrospective Benchmarking Study. Journal of Medical Internet Research. 2026;28:e101711.
5. Song J, Xu Z, He M, Feng J, Shen B. Graph retrieval augmented large language models for facial phenotype associated rare genetic disease. NPJ digital medicine. 2025;8(1):543.
6. Song J, Feng J, Zhang Y, Bi C, Zheng X, Xu Z, et al. Augmenting large language models with clinical knowledge graph for personalized perioperative fluid therapy question answering. PLOS Digital Health. 2026;5(6):e0001474.
7. Gargano MA, Matentzoglu N, Coleman B, Addo-Lartey EB, Anagnostopoulos AV, Anderton J, et al. The Human Phenotype Ontology in 2024: phenotypes around the world. Nucleic acids research. 2024;52(D1):D1333-D46.
8. Lobo M, Lamurias A, Couto FM. Identifying human phenotype terms by combining machine learning and validation rules. BioMed Research International. 2017;2017(1):8565739.
9. Groza T, Gration D, Baynam G, Robinson PN. FastHPOCR: pragmatic, fast, and accurate concept recognition using the human phenotype ontology. Bioinformatics. 2024;40(7):btae406.
10. Luo L, Yan S, Lai P-T, Veltri D, Oler A, Xirasagar S, et al. PhenoTagger: a hybrid method for phenotype concept recognition using human phenotype ontology. Bioinformatics. 2021;37(13):1884-90.
11. Feng Y, Qi L, Tian W. PhenoBERT: a combined deep learning method for automated recognition of human phenotype ontology. IEEE/ACM Transactions on Computational Biology and Bioinformatics. 2022;20(2):1269-77.

12. Yang J, Liu C, Deng W, Wu D, Weng C, Zhou Y, et al. Enhancing phenotype recognition in clinical notes using large language models: PhenoBCBERT and PhenoGPT. Patterns. 2024;5(1).
13. Garcia BT, Westerfield L, Yelemali P, Gogate N, Rivera-Munoz EA, Du H, et al. Improving automated deep phenotyping through large language models using retrieval-augmented generation. Genome Medicine. 2025;17(1):91.
14. Agarwal D, Angell R, Monath N, McCallum A, editors. Entity linking via explicit mention-mention coreference modeling. Proceedings of the 2022 Conference of the North American Chapter of the Association for Computational Linguistics: Human Language Technologies; 2022.
15. Sung M, Jeon H, Lee J, Kang J, editors. Biomedical entity representations with synonym marginalization. Proceedings of the 58th annual meeting of the association for computational linguistics; 2020.
16. Yuan H, Yuan Z, Yu S, editors. Generative biomedical entity linking via knowledge base-guided pre-training and synonyms-aware fine-tuning. Proceedings of the 2022 Conference of the North American Chapter of the Association for Computational Linguistics: Human Language Technologies; 2022.
17. Garda S, Leser U. BELHD: improving biomedical entity linking with homonym disambiguation. Bioinformatics. 2024;40(8):btae474.
18. Song J, He M, Ren S, Shen B. An explainable dataset linking facial phenotypes and genes to rare genetic diseases. Scientific Data. 2025;12(1):634.
19. Haldeman-Englert CR, Jewett T, Mulle JG, Gambello MJ, Russo RS, Murphy MM, et al. GeneReviews®. University of Washington, Seattle Seattle (WA). 1993.
20. Weissenbacher D, Zhao X, Priestley JR, Szigety KM, Schmidt SF, O'Connor K, et al. Automatic genetic phenotype normalization from dysmorphology physical examinations: an overview of the BioCreative VIII—Task 3 competition. Database. 2025;2025:baaf051.
21. OpenAI. Vector embeddings. 2026 [cited 2026 Augest 13]; Available from: https://developers.openai.com/api/docs/guides/embeddings.
22. Gutmann M, Hyvärinen A, editors. Noise-contrastive estimation: A new estimation principle for unnormalized statistical models. Proceedings of the thirteenth international conference on artificial intelligence and statistics; 2010: JMLR Workshop and Conference Proceedings.
23. OpenAI. GPT-5.6 Sol. 2026 [cited 2026 Augest 13]; Available from: https://developers.openai.com/api/docs/models/gpt-5.6-sol.
24. Remy F, Demuynck K, Demeester T, editors. BioLORD: learning ontological representations from definitions for biomedical concepts and their textual descriptions. Findings of the Association for Computational Linguistics: EMNLP 2022; 2022.
25. Lee J, Yoon W, Kim S, Kim D, Kim S, So CH, et al. BioBERT: a pre-trained biomedical language representation model for biomedical text mining. Bioinformatics. 2020;36(4):1234-40.
26. Gu Y, Tinn R, Cheng H, Lucas M, Usuyama N, Liu X, et al. Domain-specific language model pretraining for biomedical natural language processing. ACM Transactions on Computing for Healthcare (HEALTH). 2021;3(1):1-23.
27. Huang K, Altosaar J, Ranganath R. Clinicalbert: Modeling clinical notes and predicting hospital readmission. arXiv preprint arXiv:190405342. 2019.
28. Anthropic. Introducing Claude Opus 5. 2026 [cited 2026 Augest 13]; Available from: https://www.anthropic.com/research/claude-opus-5.
29. DeepSeek. DeepSeek-V4-Pro GA Release 2026 [cited 2026 Augest 13]; Available from: https://api-docs.deepseek.com/news/news260813/.
30. Vasilevsky NA, Toro S, Matentzoglu N, Flack JE, Mullen KR, Hegde H, et al. Mondo: integrating disease terminology across communities. Genetics. 2026;232(4):iyaf215.
31. Davis AP, Wiegers TC, Rosenstein MC, Mattingly CJ. MEDIC: a practical disease vocabulary used at the Comparative Toxicogenomics Database. Database. 2012;2012:bar065.
32. Schoch CL, Ciufo S, Domrachev M, Hotton CL, Kannan S, Khovanskaya R, et al. NCBI Taxonomy: a comprehensive update on curation, resources and tools. Database. 2020;2020:baaa062.
33. Doğan RI, Leaman R, Lu Z. NCBI disease corpus: a resource for disease name recognition and concept normalization. Journal of biomedical informatics. 2014;47:1-10.
34. Pafilis E, Frankild SP, Fanini L, Faulwetter S, Pavloudi C, Vasileiadou A, et al. The SPECIES and ORGANISMS resources for fast and accurate identification of taxonomic names in text. PloS one. 2013;8(6):e65390.
35. Garda S, Weber-Genzel L, Martin R, Leser U. BELB: a biomedical entity linking benchmark. Bioinformatics. 2023;39(11). doi: 10.1093/bioinformatics/btad698.

**Supplementary Figure S1**

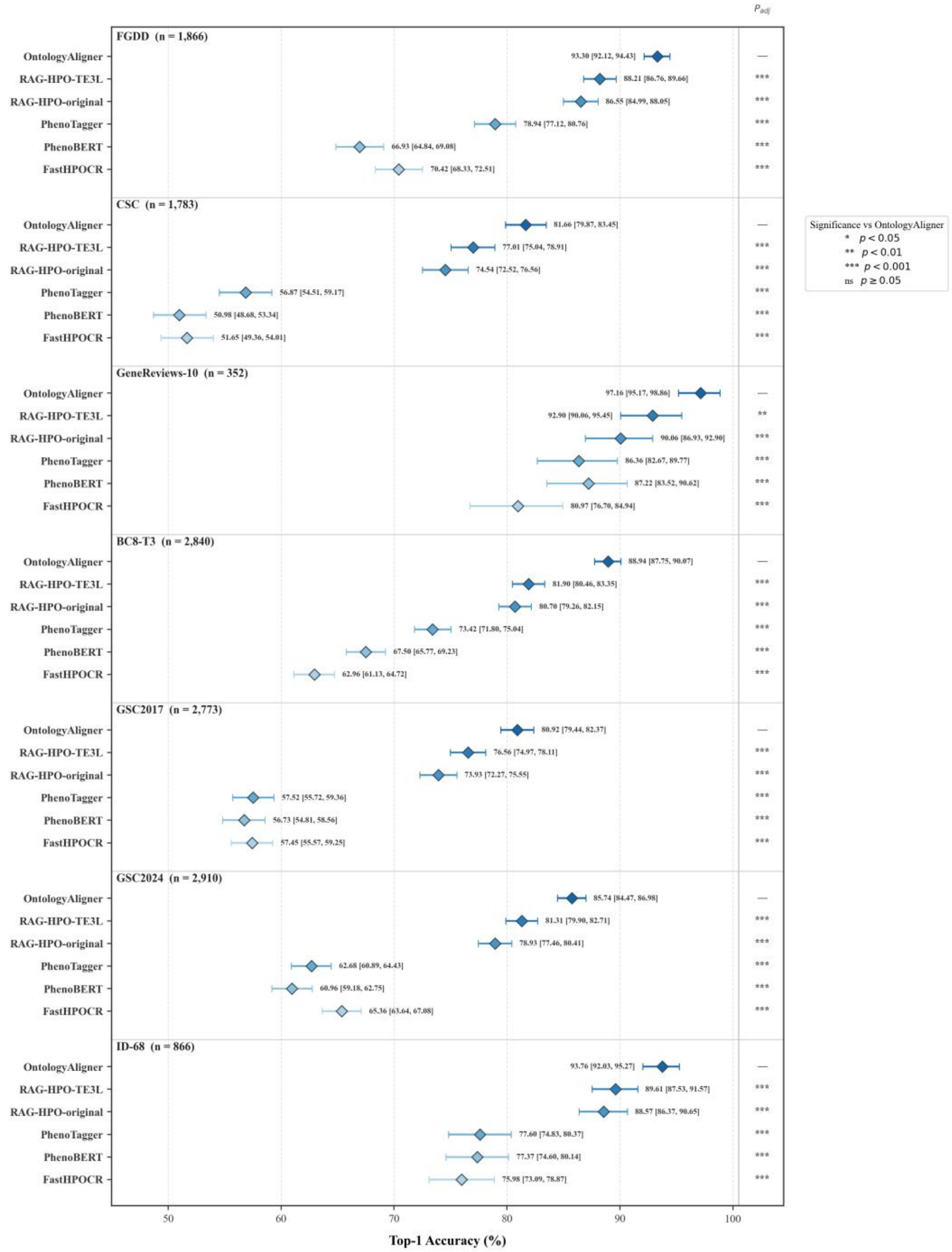

**Supplementary Figure S1. Dataset-level Top-1 accuracy across the seven constituent datasets of PhenoNormBench.** Each baseline is compared with OntologyAligner using a paired exact McNemar test, with five comparisons within each dataset adjusted using the Holm method. ** P_adj<0.01; *** P_adj<0.001.

## Supplementary Figure S2

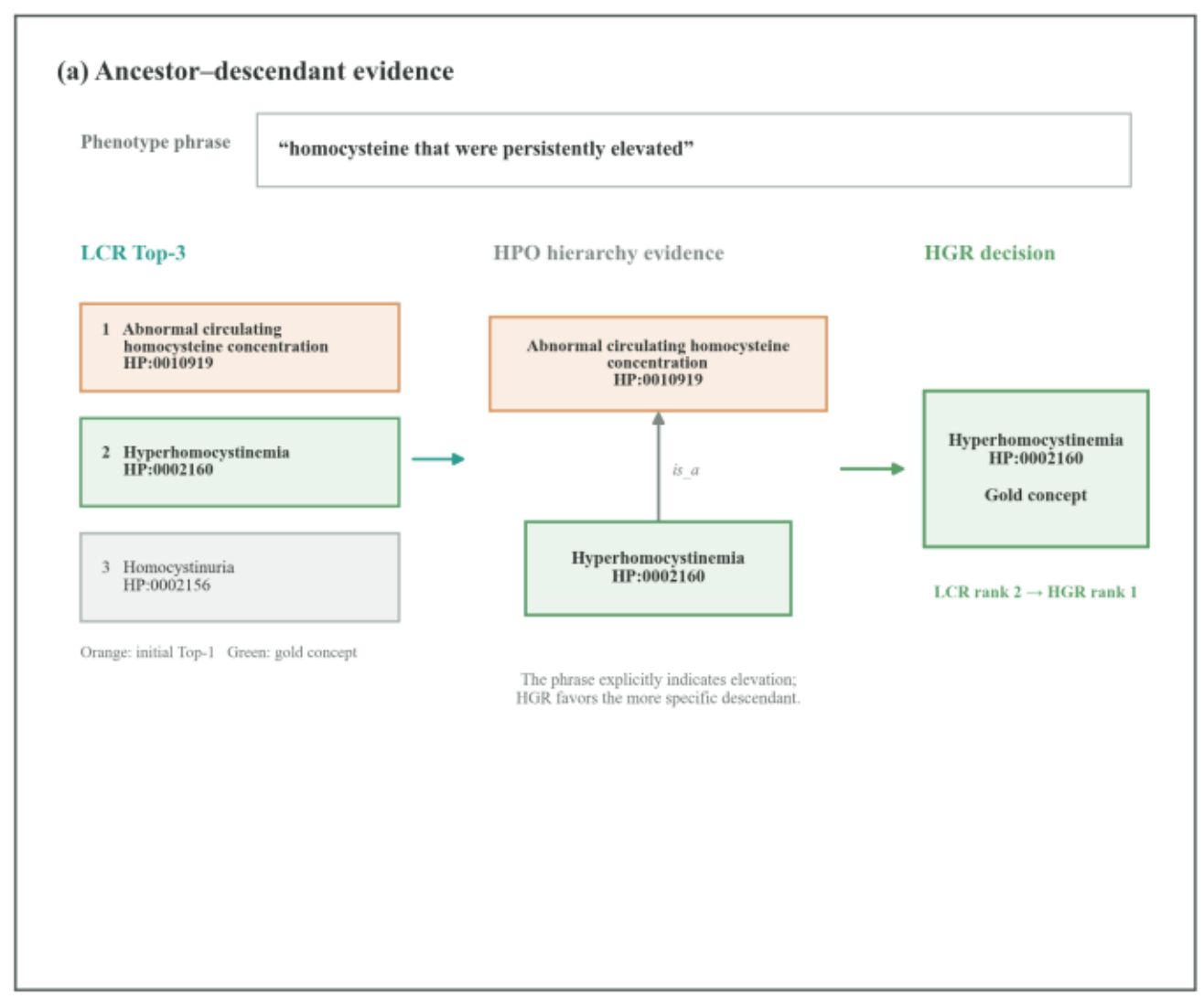

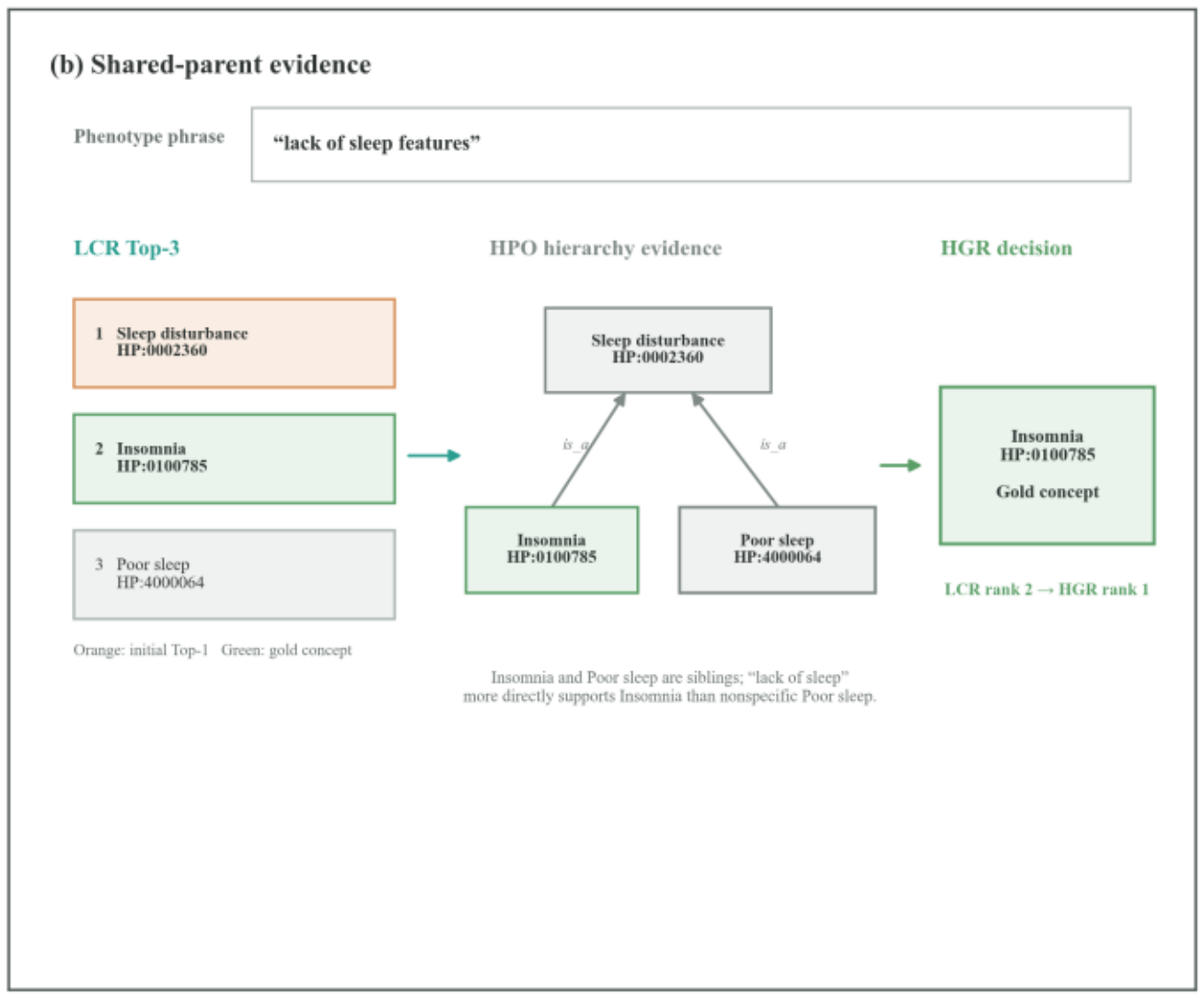

**Supplementary Figure S2. Qualitative analysis of hierarchy-guided refinement.** Two wrong-to-right examples show an parent-child case and a sibling case.

**Supplementary Table S1**

**Supplementary Table S1. LLM and embedding backbone specifications used in OntologyAligner.**

| Role | Model | Backend/provider | Exact identifier | Dimension | Official source | Access date |
|---|---|---|---|---|---|---|
| LCR/HGR | GPT-5.6-sol | OpenAI | gpt-5.6-sol | - | https://developers.openai.com/api/docs/models/gpt-5.6-sol | 2026-07-01 |
| LLM backbone ablation | Claude-Opus-5 | Anthropic | claude-opus-5 | - | https://platform.claude.com/docs/en/models/opus-5/overview | 2026-07-24 |
| LLM backbone ablation | DeepSeek V4 Flash | DeepSeek | DeepSeek-V4-Flash-0731 | - | https://api-docs.deepseek.com/quick_start/pricing/ | 2026-08-01 |
| LLM backbone ablation | DeepSeek V4 Pro | DeepSeek | DeepSeek-V4-Pro-0813 | - | https://api-docs.deepseek.com/quick_start/pricing/ | 2026-08-15 |
| OAR | TE3L | OpenAI API | text-embedding-3-large | 3,072 | https://developers.openai.com/api/docs/models/text-embedding-3-large | 2026-07-01 |
| Embedding ablation | TE3-small | OpenAI API | text-embedding-3-small | 1,536 | https://developers.openai.com/api/docs/models/text-embedding-3-small | 2026-07-01 |
| Embedding ablation | BioLORD | sentence-transformers | FremyCompany/BioLORD-2023 | 768 | https://huggingface.co/FremyCompany/BioLORD-2023 | 2026-07-01 |
| Embedding ablation | BioBERT | transformers | dmis-lab/biobert-base-cased-v1.2 | 768 | https://huggingface.co/dmis-lab/biobert-base-cased-v1.2 | 2026-07-01 |
| Embedding ablation | PubMedBERT | transformers | microsoft/BiomedNLP-BiomedBERT-base-uncased-abstract-fulltext | 768 | https://huggingface.co/microsoft/BiomedNLP-BiomedBERT-base-uncased-abstract-fulltext | 2026-07-01 |
| Embedding ablation | ClinicalBERT | transformers | emilyalsentzer/Bio_ClinicalBERT | 768 | https://huggingface.co/emilyalsentzer/Bio_ClinicalBERT | 2026-07-01 |

**Supplementary Table S2**

**Supplementary Table S2.** LLM prompt templates

| Stage | System prompt | User prompt |
|---|---|---|
| **LCR** | You are an expert biomedical ontology curator performing phenotype concept normalization to Human Phenotype Ontology.<br><br># Instructions<br>1. Rank every candidate concept from the best semantic match to the worst semantic match for the original biomedical text.<br>2. Evaluate each candidate independently using its preferred name, synonyms, and definition. Prefer exact semantic equivalence and the most specific concept that fully captures the text over broader, narrower, or merely related concepts.<br>3. Preserve distinctions such as anatomical site, clinical quality, severity, laterality, and disease versus phenotype.<br>4. Do not infer clinical details that are not stated in the original text, and do not treat the input candidate order as evidence.<br><br># Output format<br>1. If at least one candidate is a valid semantic match, return exactly one JSON object in this form:<br>{"ranking":["ID1","ID2"]}<br>2. The ranking array must contain every provided candidate ID exactly once, with no missing, duplicated, or additional IDs.<br>3. If none of the candidates is a valid semantic match, return exactly the plain text No Match instead of JSON.<br>4. Return only the JSON object or No Match. Do not include explanations, scores, comments, or Markdown. | The JSON below contains an original biomedical phrase and 20 candidate ontology concepts.<br><br>Rank all 20 candidates from best to worst according to how precisely and completely each concept represents the meaning of the original phrase.<br><br>{input_payload}<br><br>Return only the required JSON object with every candidate ID exactly once, or return exactly No Match if none is valid. |
| **HGR** | You are an expert biomedical ontology curator refining phenotype concept normalization to Human Phenotype Ontology with hierarchy evidence.<br><br># Instructions<br>1. Use the supplied HPO graph context together with preferred names, synonyms, and definitions to refine the ordering of every candidate concept.<br>2. The graph context contains explicit is_a paths for highly similar HPO concepts. Do not infer clinical details that are not stated in the original text, and do not treat the input candidate order as evidence.<br><br># Output format<br>1. Return exactly one JSON object in this form:<br>{"ranking":["ID1","ID2"]}<br>2. The ranking array must contain every provided candidate ID exactly once, with no missing, duplicated, or additional IDs.<br>3. Return JSON only. Do not include explanations, scores, comments, or Markdown. | The JSON below contains an original biomedical phrase, 20 candidate HPO concepts, and additional HPO graph context for several highly similar concepts.<br><br>Rank all 20 candidates from best to worst according to how precisely and completely each concept represents the meaning of the original phrase.<br><br>Some candidates are highly similar. Pay particular attention to the HPO graph relationships and definitions provided for these concepts, and use this information to distinguish them carefully.<br><br>{input_payload}<br><br>Return only the required JSON object. The ranking must include every candidate ID exactly once, ordered from the best match to the worst match. |